\documentclass[preprint,12pt]{elsarticle}

\usepackage{amssymb}
\usepackage{hyperref}
\usepackage{amsmath}
\usepackage{graphicx} 
\usepackage{multirow}
\usepackage{booktabs}

\usepackage{float}
\usepackage{caption}

\journal{arXiv e-prints}

\begin{document}

\begin{frontmatter}



\title{Server-side Anti-cheat in FPS games for Aimbot detection using Deep learning and Machine learning.\\}


\author[inst1]{Siddhesh A. Dhinge} 
\affiliation[inst1]{
  organization={dept. of Information Technology, Pune Institute of Computer Technology},
  city={Pune},
  country={India },
  email={dhinge.siddhesh22@gmail.com}
}

\author[inst1]{Shubham G. Sukum} 
\affiliation[inst1]{
  organization={dept. of Information Technology, Pune Institute of Computer Technology},
  city={Pune},
  country={India },
  email={shubhamsukum1722002@gmail.com}
}

\author[inst1]{Harsh S. Ranjane} 
\affiliation[inst1]{
  organization={dept. of Information Technology, Pune Institute of Computer Technology},
  city={Pune},
  country={India },
  email={harshranjane2411@gmail.com}
}

\author[inst1]{Ruturajsingh R. Rajput} 
\affiliation[inst1]{
  organization={dept. of Information Technology, Pune Institute of Computer Technology},
  city={Pune},
  country={India },
  email={ruturajsinghrajput@gmail.com}
}

\author[inst1]{Jyoti H. Jadhav} 
\affiliation[inst1]{
  organization={dept. of Information Technology, Pune Institute of Computer Technology},
  city={Pune},
  country={India },
  email={jhjadhav@pict.edu}
}

\begin{abstract}
Modern video games are becoming more complex day by day. Most of these modern games are multiplayer first-person shooter (FPS) games. The rising popularity of FPS games emphasizes the need to combat cheating for fair and enjoyable gaming.
As the number of players using cheating techniques like aimbots, wallhacks, and speed hacks is also increasing, we need a way to detect players who are using cheating tools to gain an unfair advantage over regular players. In this system, we focus exclusively on detecting aimbot cheats.
Players who use aimbot cheats generally do not prioritize other aspects of the game. To distinguish between regular and cheating players, we identify specific features encompassing time-series data such as aim velocity, number of shots, distance to target, and more, along with behavioral data such as utility usage, player movement, and other gameplay patterns. Utilizing these features, we construct a server-side aimbot detection classifier named ‘YAACS’. YAACS comprises a parser, a deep learning model, and intermediary connection utilities designed for integration with the game server.
The proposed system achieves a classification accuracy of 88.6\% with a false positive rate of 0.97\% using a Stacked LSTM with Dense layers trained on sequences of 128 ticks (Tick Delta Negative=56, Tick Delta Positive=24), outperforming the Decision Tree baseline which achieves a higher accuracy of 96.2\% but at a false positive rate of 2.68\%, 2.76x worse than the best LSTM configuration. These results demonstrate that incorporating temporal context through sequence modelling is critical for minimising false accusations in FPS cheat detection.

\end{abstract}




\begin{keyword}
First Person Shooter video game \sep Deep learning \sep Aimbot detection \sep Long Short-Term Memory (LSTM) \sep Time Series Classification \sep Cheat Detection \sep Sequence Modelling \sep Anti-cheat Systems
\end{keyword}

\end{frontmatter}



\section{Introduction}
\label{sec:introduction}
\par Video games have grown immensely in popularity and profitability among today's youngsters. Integrity and fair play are essential components of the gaming experience, especially in contemporary multiplayer games. First-person shooter (FPS) games are the genre in which the majority of these modern games belong. First-person shooter games are getting more and more popular, drawing thousands of gamers every day. At the forefront are games like Call of Duty: Modern Warfare, Valorant, Counter-Strike: Global Offensive, and many more. Balanced, skill-based encounters are essential for competitive and online first-person shooter games, allowing players to assess their skills on an even playing field.
\par Authentic and fair gameplay becomes more difficult to maintain as online gaming communities and organizations grow. Modern first-person shooter games are more complicated than ever, which causes more issues for gamers and less equitable gameplay. This is frequently the result of specific groups getting hold of game codes and abusing the title and its assets. This is when it becomes clear that an anti-cheat system is essential in the current game world.
\par The video game industry has grown rapidly in recent years as a result of developments in social connections, technology, and entertainment. The business has seen a rise in competitive esports gaming and grabbing the interest of millions with a huge worldwide audience from many cultures, regions, and styles. Due to which the popularity of playing FPS games have risen to a large extent in the gaming community. The gaming business has reached new heights because of technological advancements including amazing graphics and visuals, a real gaming experience, improved sound quality and other elements. This trend has important economic ramifications because game publishers and developers earn billions of dollars annually through esports gaming competitions, hardware sales, and game creation. The Online Multiplayer mode in esports games give players an opportunity to compete with different players around the world due to which the particular player can achieve its global rank among different players. Also the multiplayer functionality of video games promotes extensive communities and social interactions among the players around the world. Additionally, the video game industry promotes creativity and innovation.

\par Currently, the number of players using cheating techniques such as aimbots, wallhacks, and speed hacks are on the rise. Due to which the fairplay among the players in the multiplayer mode of the game is decreasing gradually. To detect these foul activities, multiple cheating detection techniques are formed including client-side anti-cheats and server-side anti-cheats. Client-side anti-cheats typically scan processes that read game’s memory or search for untrusted libraries and processes. Server-side anti-cheats, on the other hand, employ more elaborate methods, which can include log analysis or deep learning techniques for identifying cheating. Currently, we will be only focusing on detecting aimbot cheat using server-side techniques in the game Counter-Strike: Global Offensive.

\section{Literature Survey}
\label{sec:LiteratureSurvey}

\par The need for anti-cheat in video games is growing, notably in the context of FPS games, there have been many efforts made in this field.	
\par The GAN-Aimbots is a technique that was used to create an aimbot that mimics the mouse movements of a normal user and it was undetectable to most of the common anticheats. However, the results of this technique decreased and was caught when a strong aimbot was used by the cheater. Even though it was less likely to be considered suspicious by the experienced judges. The dataset that was considered for testing was of 23 players who played a few matches without cheat, a few with lightweight cheats, and the remaining with strong aimbots~\cite{b1}.

\par Cheating in games has been increasing day by day nowadays. If some of the players cheat then it affects the gaming behavior of other players in a bad way who play the game with sincerity. A statistical model was developed to understand such effect and consequences of it which concluded that continuous cheating amplify the spread of this problem and make it even more widespread and difficult to manage. This statistical model considered the dataset of about 40 servers~\cite{b8}.

\par The Dynamic Bayesian approach was used to identify cheater by considering their behavior in the game. The behavior of a normal player and a cheater will be different because the probability threshold of a cheater fluctuates while a generous player's probability remains below the threshold. This approach also checks the aiming precision of various players. The training data used for this approach consisted of 16 players and it was trained and tested by the number of players cheating and other were normal players~\cite{b10}.

\par The interview-based study was done to understand the behavior of computer gamers experience and how they consider certain conditions. Cheating also affects the gaming experience of a gamer. The experiences in first-person shooter games are influenced by their goals and the variety in the game. If the goals and requirements of the gamer are not satisfied then it may have a bad impression on the gamer.This study considered over 12 people who were interviewed~\cite{b12}.

\par Datasets used SWAT(secure water treatment), WADI(water distribution), MSL(mars science laboratory rover). RNN based model shows superiority in terms of temporal dependencies but RNN restricts parallelization because it computes its output sequentially whereas transformer takes  one sequence one at a time so parallel processing is possible and GNN based model provides results better than the others because dimensions of the dataset increases~\cite{b2}.

\par Dataset was collected using log files which were generated during gameplay and from the log files features were extracted such as hits taken, hits given, damage dealt and damage taken. Decision tree is highly dependent on individual player configuration which also leads to various numbers of false positives so it can only be used as a classifier. The naive bayes classifier was only applicable to single behavior dependency on categorical data. The keystroke dynamics based SVM’s had a wide array of accuracy ratings. This could be because of the huge numbers of features being considered~\cite{b3}.

\par Dataset used here was collected in real world context of players from the game Counter-Strike: Global Offensive. Results show that the model was able to clearly distinguish between fair and fraudulent gameplay. In each domain, a new hyperparameter search should be performed to find good functioning architecture. Cross validation was able to show that the model learnt things that are not player specific. Model detected unfair play even when they weren’t exposed to their behavior~\cite{b4}.

\par To detect aimbot machine learning classification techniques such as Support Vector Machine were used. To classify excellent players from cheaters behavioral data was used. It is determined that players who cheat have different behavioral patterns than genuine players. First classification is done to identify cheating and excellent players then another classification is used to classify cheating players from excellent players. The authors created a system named Aimdetect for Counter-Strike 1.6 and Counter-Strike: Global Offensive. The game servers were hosted publicly for others to join and play. Through this, they were able to maintain a low false positive rate as small as 0.7\%~\cite{b14}.

\par Instead of using complete time series data for detecting cheating players the time series data is divided into frames of 60 seconds. Then statistical methods are applied to those 60 seconds of data to create features like the Number of Records, Number of Visible-Target Rows, Visible-To-Total Ratio, and more. These features are then fed to 2 classification models - Logistic Regression and Support Vector Machine. The system was able to achieve a high prediction accuracy of nearly 90\%~\cite{b9}.

\par The authors discussed two new features - Time on Target (ToT) and Cursor Acceleration when acquiring aim (AccA). It was discovered that ToT was a better feature than AccA. Statistical method two-sample Kolmogorov-Smirnov tests were used to identify differences between honest and cheating players~\cite{b13}.

\par To detect Wall hack cheats four new features were discussed - Frequency of illegal traces, consecutively of illegal traces, Distance to world traces, Distance to entity traces. The intuition behind these features was that players who are using visual-based cheats don't see entities such as a wall or other obstacles and have different behaviors near these entities. The paper then discussed that the system was able to provide an accuracy of 70\%~\cite{b11}.

\par Detecting cheats in FPS games through visual analysis involves using advanced techniques like Deep Neural Networks (DNNs). These networks are trained on a bunch of frames containing both cheating and fair gameplay instances. By doing this, the DNN learns to distinguish between the two by looking at visual cues in the frames. The best DNN architecture for this task combines both global and local features. Although the DNN works really well, it doesn't give exact numbers for metrics like sensitivity and precision. The visual cheats are classified in 3 categories minimal, medium and full each of which provides a different level of visual assistance. The dataset used for training includes frames from two popular shooter games, referred to as Game1 and Game2 in the paper~\cite{b5}.

\par When looking at ways to stop cheating in video games, various methods can be used. These methods are compared across different categories like how well they handle tampering, how easy they are to put into the game, how much they slow down the game, how invasive they are, and how well they work with different types of games. There are two main types: ones that work on the player's computer (client-side) and ones that work on the game's server (server-side). Server-side methods are good at stopping cheating but can make the game slow because they use a lot of resources. Client-side methods are lighter on resources, but a technically skilled person (cheater) might find ways to get around them more easily~\cite{b6}.

\par There's this new thing called TSViz that helps us understand deep learning models, especially convolutional deep learning (CNN) models. With TSViz, we can explore and analyze the network in different ways and at different levels. It helps us figure out which parts of the network are making specific predictions and why certain filters in the network are important. Plus, it helps us see how well the network can handle noisy data that might mess up its predictions. They used an Internet Traffic Dataset to train the model for time-series forecasting, with inputs consisting of 50 time-steps. They trained it using small batches of 5 till 5000 epochs~\cite{b7}.

\par In ~\cite{b15} proposed a generic system architecture for aimbot detection in FPS games, outlining potential deep learning and statistical approaches for tackling the problem. The present study builds directly upon that foundational architecture, extending it with a concrete implementation and systematic experimentation, evaluating 9 LSTM model configurations across 6 dataset configurations with varying temporal context windows, and introducing a comparative analysis against a Decision Tree baseline with particular emphasis on minimising false positive rates for production anti-cheat deployment.

\section{System Architecture}
\label{sec:SystemArchitecture}
\par Aimbot cheats make it much easier for players to dominate. Because of this, players that utilize aimbot cheats often don't give other aspects of the game priority. This may include avoiding most of the tools provided by the game itself or not being affected by the movement inaccuracy which other individuals might have experienced when shooting. A few significant traits that are useful in recognizing anomalous plays include the behavioral data paired with the time series data that is generated by the players, such as player placement and the rate at which they react during battles. These characteristics, including time series data and player behavioral information, will be discovered in an attempt to distinguish between honest and dishonest individuals. We build a server-side aimbot detection classifier called "YAACS" (Yet Another Anti Cheat System) incorporating these features. A parser, a model developed using deep learning, and intermediate concentrations connection tools aimed for interaction with the game server will be included in this system. \\

\begin{figure}[htp]
    \centering
    \includegraphics[width=10cm]{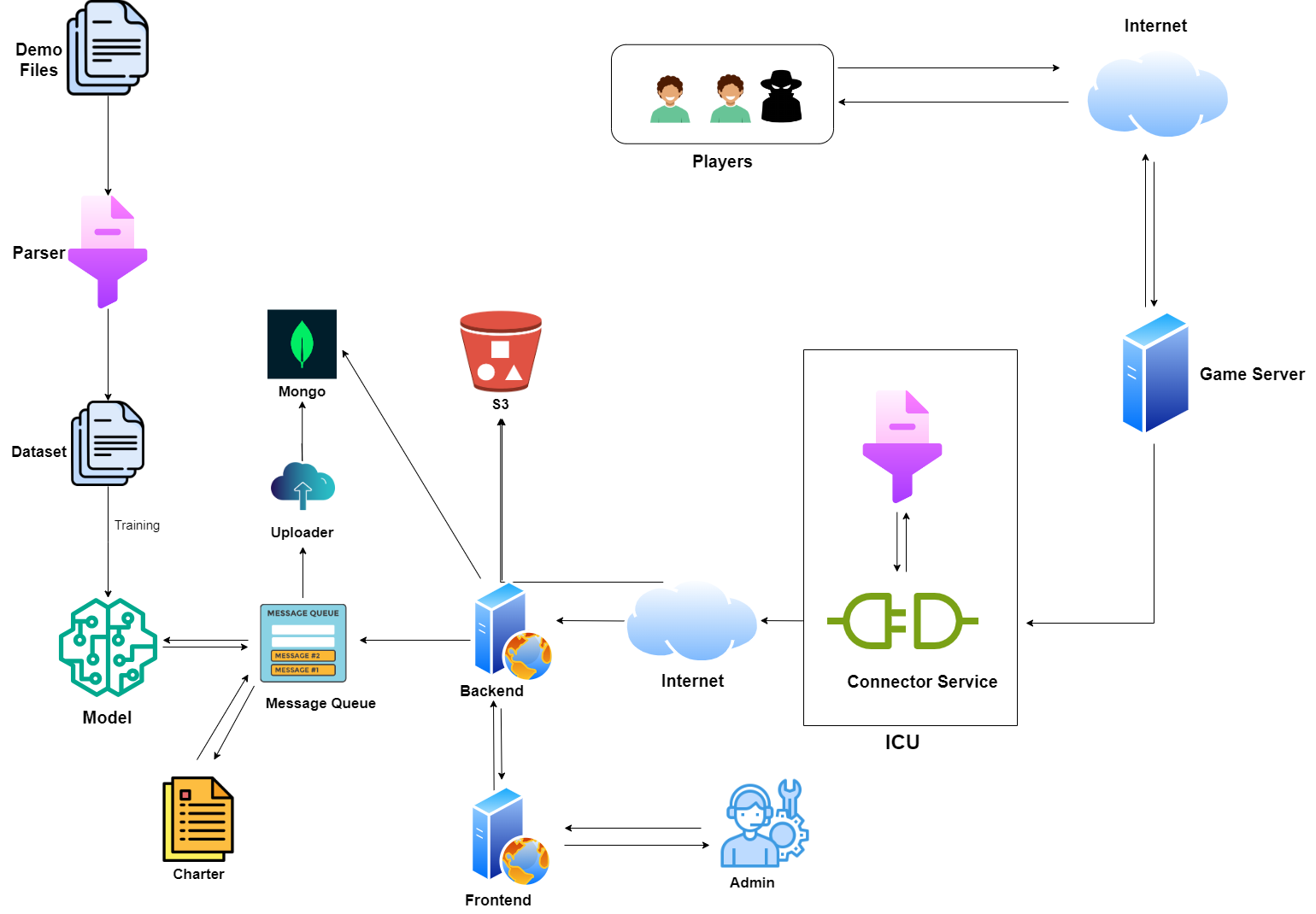}
    \caption{System Architecture}
    \label{fig:architecture}
\end{figure}
\subsection{Demo Files}

\par
    The Source recorder is employed to create demo files, which include recorded events that may be viewed and replayed in-game. Playing back demo files takes you back to that specific tick in the game's history. In the game, a tick is a particular point in time. Professional Counter-Strike matches tend to be played at a 128 tickrate, which means that 128 game state modifications occur in a second between the client and server. The tick rate is an indicator of how often the server updates its state. Additionally, the demonstrations are captured at a distinct pace known as the snapshot rate. The number of changes written to the demo file each second is determined by the snapshot rate. Usually, this is adjusted lower to limit bandwidth use to 32 or 64 ticks. Demos with a snapshot duration of 128 ticks and a tick rate of 128 ticks will be taken into consideration for our system as they will precisely offer game state information.Usually, this is adjusted lower to limit bandwidth use to 32 or 64 ticks. Demos with a snapshot duration of 128 ticks and a tick rate of 128 ticks will be taken into consideration for our system as they will precisely offer game state information. \\

\subsection{Parsing, Feature Engineering, and data cleaning}
\par
The format of these demo files is different from what we require; therefore, a parser that can correctly parse such demo files must be used to resolve this problem. We are going to use a Python-demo parser for our system. This tool can provide us with access to the game events that we can monitor. Features such as player position, yaw, pitch, current tick, velocity between ticks, and more will be the main focus of our system. The deep learning model would then use these features after they have been cleaned, normalized, and standardized. \\

\subsection{Game Server}
\par
The game server holds the actual game state. Clients fetch and update this game state as the game progresses. As the game is multiplayer this adds latency to the clients. In CounterStrike: Global Offensive latency is measured as the roundtrip time required for data to be transferred between client and server. This means if the game server is on time 150ms and the clients are on a latency of 50ms then the client state is at 125ms. To minimize this latency clients generally include mechanisms such as interpolation which eases out these effects. \\

\subsection{Intermediatory Connection Utilities}
\par
Events are recorded to the demo file and players are linked to the game server. Our model requires to receive the parsed demo file from the game server in some way. We can accomplish this by using a script to determine whether the current match is finished. Following that, we move the match's parsed demo file from the game server to the web server. \\

\subsection{Web Server (Frontend and Backend)}
\par
It displays all the names of the players in the match, shows which of them were cheating, and displays reports regarding their movements, activity, results, etc. Frontend’s access is given to the admin of our Anti-cheat system. Backend ensures the website performs correctly, focusing on databases, back-end logic, and application programming interface (APIs) \\

\subsection{Messaging Queue}
\par
Facilitates asynchronous communication between components of the YAA-C system, enabling seamless coordination and scalability in processing tasks and data flows.
\\

\subsection{Charter}
\par
Performs the task of creating all the analysis charts and diagrams from the match data obtained for post-match analysis.
\\

\subsection{Uploader}
\par
Performs the task of uploading data from the charter and model to the Mongo.
\\

\subsection{Mongo}
\par
It facilitates storing data of all analysis charts created and storing results of the model.
\\

\subsection{S3}
\par
S3 is used for storing the analyzed files for future analysis and improvement in the system.
\\

\subsection{Deep Learning Model}
\par
LSTM, RNN and decision tree will be the model we choose to distinguish between honest and dishonest players. RNNs meet our features well because they can be applied to time-series data and we are choosing decision tree to compare the results.
\\

\section{Methodology}
\label{sec:Methodology}

\subsection{Feature Identification}

\par We need to track multiple features from the game server so that we can build an effective model. All the features passed from the game server are categorized into 2 types:

\begin{itemize}
    \item \textbf{Features used in model}: These features are directly fed to the model we created. We will be further referring to these features as \textit{alpha features}.
    
    \item \textbf{Features used for monitoring and reporting}: These features are primarily used for player identification, monitoring, and post-match report generation. We are calling these features \textit{beta features}.
    
\end{itemize}

\subsection{Feature Extraction and Engineering}

\par We will be extracting the following features through our custom-built parser from the demo files provided by the game server.

\vspace{0.3cm}

\begin{enumerate}
    \item \textbf{Current Tick (beta feature):} Denotes the current fight tick, initiating at 0 and advancing progressively. For instance, if there are ticks at (10, 11, 12, 14, 15), and the tick at 13 is missing, the new sequence of ticks will be (0, 1, 2, 4, 5). The Current Tick is computed as:
    \[ \text{Current Tick} = \text{tick} - \text{Min Tick} \]
    
    \item \textbf{Player Id (beta feature):} Represents the identifier of the player currently under tracking.\\
    
    \item \textbf{Player Name (beta feature):} Indicates the name of the player currently being tracked.\\

    \item \textbf{Player Position (beta feature):} 
    \begin{itemize}
        \item \textbf{X:} Current X coordinate of player.
        \item \textbf{Y:} Current Y coordinate of player.
        \item \textbf{Z:} Current Z coordinate of player.\\
    \end{itemize}
    
    \item \textbf{Delta Position (alpha feature):} 
    Difference between the previous position of the player and the current position of the player.
    \begin{itemize}
        \item \(\Delta X = X_{\text{currentTick}} - X_{\text{previousTick}}\)
        \item \(\Delta Y = Y_{\text{currentTick}} - Y_{\text{previousTick}}\)
        \item \(\Delta Z = Z_{\text{currentTick}} - Z_{\text{previousTick}}\)\\
    \end{itemize}

    \item \textbf{View Angles (beta feature):} 
   Refers to the view angles of tracked players, excluding roll as it is typically unused in FPS games.
    \begin{itemize}
        \item \textbf{Yaw:} Represents the horizontal direction where the player is currently looking, ranging from 0 to 359 degrees. For example, if the player is currently facing a yaw of 3 degrees and begins to move the mouse to the right, the subsequent yaws will follow a sequence such as $[2, 1, 0, 359, 358, 357, \ldots]$.
        
        \item \textbf{Pitch:} Represents the vertical direction where the player is currently looking, with values ranging from $-90$ to $90$ degrees. For example, if the player is currently looking at a pitch of $-6$ degrees and begins moving the mouse downward, the subsequent pitches will follow a sequence of angles which will form \\
        $[-5, -4, -3, -2, -1, 0, 1, 2, 3, \ldots]$.\\
    \end{itemize}
    
    \item \textbf{Delta View Angles (alpha feature):} 
    Difference between the player's previous and current view angles, consisting of:
    \begin{itemize}
        \item \(\Delta\text{Yaw}\): The difference between the player's previous and current yaw angles.
        \item \(\Delta\text{Pitch}\): The difference between the player's previous and current pitch angles.\\
    \end{itemize}

    \item \textbf{Delta Aim Arc (alpha feature):} Represents the length of the arc formed in 3D space with \( R \) as the distance between the player and the target, and two spherical angles \(\{\text{yaw}_{\text{previous tick}}, \text{pitch}_{\text{previous tick}}\}\), \(\{\text{yaw}_{\text{current tick}}, \text{pitch}_{\text{current tick}}\}\) after converting to space vectors.\\
    
    \item \textbf{Is Flashed (alpha feature):} Indicates whether the player is currently flashed. The values are between 0 and 1, where 0 represents that the player is not flashed at all, and 1 represents that the player is fully flashed. Intermediate values are interpolated using a bezier curve.\\
    
    \item \textbf{Is Crouching (alpha feature):} Indicates whether the player is currently crouching. A value of 1 indicates that the player is in between standing and crouching, a value of 2 indicates that the player is completely crouched, and a value of 0 indicates that the player is standing.\\
    
    \item \textbf{Is In Air (alpha feature):} Indicates whether the player is currently in the air and not touching the ground. True represents that the player is in the air, while False represents that the player is on the ground.\\
    
    \item \textbf{Utility Damage Done (alpha feature):} Represents the amount of utility damage done until this tick (inclusive), including HE grenades and Molotov cocktails as the damaging utility.\\
    
    \item \textbf{Support Utility Used (alpha feature):} Represents the number of utility supports done until this tick (inclusive), including flashbang and smoke grenades as support utilities.\\
    
    \item \textbf{KDR (alpha feature):} Represents the Kill-Death Ratio of the player.\\
    
    \item \textbf{Is Firing (alpha feature):} Indicates whether the player is currently shooting the current weapon.\\
    
    \item \textbf{Target Id (beta feature):} Id of the player who is currently being attacked in fight.\\
    
    \item \textbf{Target Name (beta feature):} Name of the player who is currently being attacked in fight.\\
    
    \item \textbf{Delta View Angles (alpha feature):} 
    Difference between the player's previous and current view angles, consisting of:
    \begin{itemize}
        \item \(\Delta\text{Yaw}\): The difference between the player's previous and current yaw angles.
        \item \(\Delta\text{Pitch}\): The difference between the player's previous and current pitch angles.\\
    \end{itemize}
    
    \item \textbf{Target Position (beta feature):} 
    \begin{itemize}
        \item \textbf{targetX:} Represents the current X coordinate of the target player.
        \item \textbf{targetY:} Represents the current Y coordinate of the target player.
        \item \textbf{targetZ:} Represents the current Z coordinate of the target player.\\
    \end{itemize}
    
    \item \textbf{Target Delta Position (alpha feature):} 
    \begin{itemize}
        \item \textbf{targetDeltaX:} Represents the difference between the current tick's X coordinate and the previous tick's X coordinate for the target player.
        \item \textbf{targetDeltaY:} Represents the difference between the current tick's Y coordinate and the previous tick's Y coordinate for the target player.
        \item \textbf{targetDeltaZ:} Represents the difference between the current tick's Z coordinate and the previous tick's Z coordinate for the target player.\\
    \end{itemize}

    \item \textbf{Damage Done (alpha feature):} Represents the total damage done to the enemy, including armor.\\
    
    \item \textbf{Distance To Target (alpha feature):} Represents the distance of the player to the target in game units, where 1 game unit equals 1 inch.\\

    \item \textbf{Target Hit Area (alpha feature):} On which body part the target player was shot at. Possible values include:
    \begin{itemize}
        \item HITGROUP\_GENERIC: This denotes explosion damage.
        \item HITGROUP\_HEAD: This denotes damage to the player's head.
        \item HITGROUP\_CHEST: This denotes damage to the player's chest.
        \item HITGROUP\_STOMACH: This denotes damage to the player's stomach.
        \item HITGROUP\_LEFTARM: This denotes damage to the player's leftarm.
        \item HITGROUP\_RIGHTARM: This denotes damage to the player's rightarm.
        \item HITGROUP\_LEFTLEG: This denotes damage to the player's leftleg.
        \item HITGROUP\_RIGHTLEG: This denotes damage to the player's rightleg.
        \item HITGROUP\_GEAR: This denotes damage to the player's gear.\\
    \end{itemize}
    
    \item \textbf{Weapon Used (alpha feature):} Which weapon was used in "weapon\_weaponName" format.\\
    
    \item \textbf{Weapon Category (alpha feature):} Represents the weapon's category. Possible categories include:
    \begin{itemize}
        \item weapon\_category\_pistol
        \item weapon\_category\_smg
        \item weapon\_category\_ar
        \item weapon\_category\_sniper
        \item weapon\_category\_shotgun
        \item weapon\_category\_lmg\\
    \end{itemize}

    \item \textbf{Is Scoping (alpha feature):} Indicates whether the player is currently using the scope of the weapon. Returns True if scoping, otherwise False.\\
    
    \item \textbf{Is Target Blind (alpha feature):} Indicates whether the target player is currently flashed. The values are between 0 and 1, where 0 represents that the target player is not flashed at all, and 1 represents that the target player is fully flashed. Intermediate values are interpolated using the bezier curve.\\
    
    \item \textbf{Shot Target Through Smoke (alpha feature):} Indicates whether the target has been shot through smoke. Either True or False. This feature is difficult to capture as the problem is to check whether 2 points in 3D are inside of an Capsule.\\
    
    \item \textbf{Target Returned Damage (alpha feature):} Represents the damage returned to the player by the target player in this round.\\

\end{enumerate}

\subsection{Data Preprocessing and Parsing}
\par The collected demo files from both data sources need to be parsed. This parsing is done to extract useful features from the demo files. The demo file is a compressed format for storing game data efficiently. To parse these demo files we will be using our Python parser for parsing demos. This Python parser is a wrapper around a Rust parser publicly available (Python-demoparser), thus it will be significantly faster and memory efficient.

\subsection{Dataset Configurations}
\par Using our created parser we parsed all the demo files. We created 6 total datasets from our demo files with different configurations. The configurations listed are:

\begin{itemize}
    \item \textbf{Tick Delta Negative}: How many ticks before a hurt event shall be considered for parsing?
    \item \textbf{Tick Delta Positive}: How many ticks after a hurt event shall be considered for parsing?
    \item \textbf{Max Fight Size}: How many ticks in total the fight sequence should compromise after the hurt events are merged? Fight sequences that are larger than this are broken down into multiple sequences.
\end{itemize}

\begin{table}[h]
\centering
\resizebox{\textwidth}{!}{
\begin{tabular}{|l|c|c|c|c|c|c|}
\hline
 & Dataset 1 & Dataset 2 & Dataset 3 & Dataset 4 & Dataset 5 & Dataset 6 \\
\hline
Tick Delta Negative & 56 & 48 & 32 & 16 & 8 & 0 \\
\hline
Tick Delta Positive & 24 & 24 & 16 & 8 & 4 & 0 \\
\hline
Max Fight Size & 128 & 96 & 64 & 32 & 16 & 1 \\
\hline
\end{tabular}
}
\caption{Dataset Configuration}
\end{table}

\subsection{Model Creation}
\par  There are 3 types of models which we have trained namely \textit{Single LSTM, Stacked LSTM, Stacked LSTM with Dense layers} and in each of the model types there are 3 different models with varying configurations from each other. A detailed description of the Model configuration of every model is given as follows- 
\newline

\subsubsection{Single LSTM}
\begin{enumerate}
    \item \textbf{Model 1:} The model starts with an LSTM (Long Short-Term Memory) layer configured with 32 units. This layer processes input data shaped to match the input sequence dimensions given by the dataset samples. Following the LSTM layer, a dense layer with one neuron and a sigmoid activation function is added to produce a binary classification output.

    \item \textbf{Model 2:} This model consists of an LSTM (Long Short-Term Memory) layer with 64 units, which processes sequential input data. The input shape is specified by the dataset samples, indicating the dimensions of the input data that the LSTM layer will accept. Following the LSTM layer, a dense layer with one neuron and a sigmoid activation function is added.

    \item \textbf{Model 3:} The model architecture begins with an LSTM (Long Short-Term Memory) layer configured with 128 units, which is suitable for processing sequential input data. The specified input shape given by the dataset samples indicates the dimensions of the input data that the LSTM layer will handle. Following the LSTM layer, a dense layer is added with a single neuron and a sigmoid activation function. 
\end{enumerate}

\subsubsection{Stacked LSTM}
\begin{enumerate}
    \item \textbf{Model 1:} The model is a Sequential neural network architecture comprising two LSTM layers with 32 units each. The input shape is determined by the dimensions of the dataset samples, where sample length represents the number of time steps and sample width denotes the number of features. The first LSTM layer returns sequences, while the second does not. Dropout layers with a rate of 1/10 are applied after each LSTM layer to mitigate overfitting. Finally, a Dense layer with a sigmoid activation function outputs a single value.

    \item \textbf{Model 2:} The model is a Sequential neural network architecture with two LSTM layers, each consisting of 64 units. The input shape is defined by the dimensions of the dataset samples, where sample length represents the number of time steps and sample width denotes the number of features. Dropout layers with a rate of 1/10 are applied after each LSTM layer to prevent overfitting. The final layer is a Dense layer with a sigmoid activation function, outputting a single value.

    \item \textbf{Model 3:} The model is structured as a Sequential neural network, featuring two LSTM layers, each with 128 units. The input shape is determined by the dimensions of the dataset samples, where sample length signifies the number of time steps, and sample width indicates the number of features. Dropout layers with a rate of 1/10 are inserted after each LSTM layer to prevent overfitting. Finally, a Dense layer with a sigmoid activation function is utilized to produce a single output value.
\end{enumerate}

\subsubsection{Stacked LSTM with Dense layers}

\begin{enumerate}
    \item \textbf{Model 1:} The Sequential model consists of two LSTM layers with 32 units each, followed by dropout layers with a rate of 1/10 to mitigate overfitting. Two Dense layers follow, the first with ReLU activation and units equal to sample length, and the second with half the number of units. Another dropout layer with the same rate is included, and the model concludes with a dense layer utilizing a sigmoid activation function, outputting a single value.

    \item \textbf{Model 2:} It begins with an LSTM (Long Short-Term Memory) layer with 64 units and an input shape corresponding to the dimensions of the input dataset (sample length, sample width). The parameter ‘return sequences=True' indicates that this LSTM layer will return sequences rather than a single output, which is useful for sequence-to-sequence models. Next, another LSTM layer with 64 units is added without return sequences=True, this layer will output a single vector for the entire sequence. Finally, the model ends with a dense layer containing a single neuron and a sigmoid activation function.

    \item \textbf{Model 3:} The model comprises two LSTM layers, each with 128 units, followed by dropout layers with a rate of 1/10 to prevent overfitting. Subsequently, two dense layers are included, the first with ReLU activation and a number of units equal to sample length, and the second with half the number of units. Another dropout layer with the same rate follows. Finally, a dense layer with a sigmoid activation function outputs a single value.
\end{enumerate}

\subsection{Training of LSTM and Decision Tree Model}
\par Once the dataset is prepared we are using the deep learning model LSTM and Decision Tree model. The training is done over GPU with 20 epochs. We then evaluated the models on the test dataset to assess their performance. We used the trained model to make predictions on new data.

\subsection{Creating Intermediatory Connection Utilities}
\par To ensure effective communication between YAACS and the game server, intermediary connection utilities (ICU) have been developed. These utilities are used to connect the proprietary game server with our YAA-C system. 

\subsection{Creating Charter}
\par Once the created dataset from the parser is passed to YAA-CS charter is given the job of creating the relevant reports for post-match analysis results. These reports are charts answering various questions about the match. There are a total of 10 reports hand-picked by us which are-

\begin{itemize}
    \item Most used Weapon in the match.
    \item Mostly targeted body parts in the match.
    \item Most used Weapon Category in the match.
    \item Movement preferred during fights in the match.
    \item When the Player could not see but shot landed on target in the match.
    \item Where do players see vertically in the match.
    \item Average Utility used in the match.
    \item Do players play with a scope on or off for ARs in the match.
    \item Do players play with scope on or off for SNIPERs in the match.
    \item Was the Player blind during fights in the match.
\end{itemize}

\subsection{Creating Uploader}
\par Once the dataset is processed by Model and Charter the match data is given to Uploader as a job. The Uploader connects with the database to update and create the required data.

\subsection{Setting up S3, MongoDB and RabbitMQ}
\begin{itemize}
    \item We are using S3 for dataset storage where we are storing the datasets after parsing by the parser.
    \item We are using MongoDB for storing the admin user details, matches data, player data, and reports.
\end{itemize}

\section{Experimentation Setup}

\par Experimental setup includes the game server connected to the deep learning model using ICU, messaging queues and other components. The deep learning model will be first trained on the training dataset. Once a client connects to the game server and plays the game as usual, the deep learning model in the background will watch over the player's actions and decisions. If the client then performs some cheating action it will be detected by the deep learning model and the server will be notified of the client's misconduct.

  \subsection{Setting of Hyperparameters}
\par There are hyperparameters which can be set when the dataset is being created. These can be used to pass fights of various sequence length to model.
\par All models created themself were created to test these different hyperparametes like number of memory cells, number of layers in LSTM, and more. 

\subsection{Module Wise Setup \\}

\subsubsection{Module Installation}
\par Each component in YAA-C can be installed with either-
\begin{itemize}
    \item Python: install script provided in source code
    \item JavaScript: Thorugh npm install.
\end{itemize}

\subsubsection{Experiment Process}
\par Once installed the typical experiment flow is mentioned below.

\begin{enumerate}
    \item \textbf{The YAA-C system is connected with the game server:} This connection is done via installation and configuration of ICU and via an API key of YAA-CS.\\
    
    \item \textbf{The client connects:} Many game clients connects to the game server. Some of these game clients are modified by using some type of cheating software to achieve aimbots.\\
    
    \item \textbf{The Match Starts:} The game clients play the FPS game and some game clients use aimbots for their advantages.\\
    
    \item \textbf{The Match Ends:} Once the match ends the game server emits the recorded game state in format of demo file. ICU captures this demo file and using parser converts it to a proper format. These parsed file is then sent to the game server.\\
    
    \item \textbf{Backend triggered:} Once the match upload endpoint is hit on backend it sends the file to the S3 for storing and posts a job to RabbitMQ for processing of these match.\\
    
    \item \textbf{Charter triggered:} Charter receives the job from RabbitMQ and starts processing it for report generation. Once the report values are rendered they are sent back to RabbitMQ as a job.\\
    
    \item \textbf{Model triggered:} Model receives the job from RabbitMQ and starts processing it for fight sequence classification. Once the model completes classification on fighting sequences the data is sent back to RabbitMQ as a job.\\
    
    \item \textbf{Uploader triggered:} Uploader receives the job from RabbitMQ and uploads it to appropriate place in the database.\\
    
    \item \textbf{Admin post-match analysis:} Now the game server admin can view the analyzed match through the frontend UI, There we are showing them reports, fight sequences and player classifications.
\end{enumerate}

\section{Results of Experiments}

\vspace{0.6cm}

\par We have initially 6 different datasets according to some parameters. Each of the datasets is focused on creating different-sized sequences so we can measure what sequence length fits our needs. For each of the datasets, three types of LSTM models are created namely \textbf{Single LSTM}, \textbf{Stacked LSTM}, \textbf{Stacked LSTM with Dense layer}, and in each of the LSTM models there are 3 unique models created with different model configurations. So results of all the models for every dataset are mentioned below:
\newline
\vspace{0.6cm}

\begin{table}[h]
\centering
\resizebox{\textwidth}{!}{
\begin{tabular}{|l|c|c|c|c|c|c|c|c|c|}
\hline

\multirow{2}{*}{Metrics}
& \multicolumn{3}{|c|}{Single LSTM}
& \multicolumn{3}{|c|}{Stacked LSTM}
& \multicolumn{3}{|c|}{Stacked LSTM with Dense layers} \\
\cline{2-10}

& Model 1 & Model 2 & Model 3
& Model 1 & Model 2 & Model 3
& Model 1 & Model 2 & Model 3 \\
\hline

Accuracy & 0.887 & 0.887 & 0.887 & 0.887 & 0.886 & 0.888 & 0.886 & 0.888 & 0.887 \\
\hline
Precision & 0.931 & 0.931 & 0.926 & 0.931 & 0.917 & 0.945 & 0.958 & 0.943 & 0.956 \\
\hline
F1 score & 0.753 & 0.753 & 0.753 & 0.753 & 0.753 & 0.753 & 0.744 & 0.753 & 0.747 \\
\hline
Recall & 0.632 & 0.632 & 0.635 & 0.632 & 0.639 & 0.626 & 0.607 & 0.627 & 0.613 \\
\hline
Loss & 0.317 & 0.312 & 0.312 & 0.311 & 0.311 & 0.310 & 0.315 & 0.315 & 0.320 \\
\hline
True Positive & 1216 & 1216 & 1222 & 1216 & 1229 & 1205 & 1169 & 1207 & 1180 \\
\hline
True Negative & 5052 & 5052 & 5045 & 5052 & 5032 & 5073 & 5092 & 5070 & 5088 \\
\hline
False Positive & 90 & 90 & 97 & 90 & 110 & 69 & 50 & 72 & 54 \\
\hline
False Negative & 707 & 707 & 701 & 707 & 694 & 718 & 754 & 716 & 743 \\
\hline

\end{tabular}
}
\caption{Dataset 1}
\end{table}

\vspace{1cm}

\begin{table}[h]
\centering
\resizebox{\textwidth}{!}{
\begin{tabular}{|l|c|c|c|c|c|c|c|c|c|}
\hline

\multirow{2}{*}{Metrics}
& \multicolumn{3}{|c|}{Single LSTM}
& \multicolumn{3}{|c|}{Stacked LSTM}
& \multicolumn{3}{|c|}{Stacked LSTM with Dense layers} \\
\cline{2-10}

& Model 1 & Model 2 & Model 3
& Model 1 & Model 2 & Model 3
& Model 1 & Model 2 & Model 3 \\
\hline

Accuracy & 0.897 & 0.897 & 0.898 & 0.897 & 0.896 & 0.897 & 0.898 & 0.899 & 0.897 \\
\hline
Precision & 0.921 & 0.917 & 0.913 & 0.921 & 0.926 & 0.920 & 0.913 & 0.912 & 0.883 \\
\hline
F1 score & 0.751 & 0.752 & 0.756 & 0.751 & 0.748 & 0.751 & 0.755 & 0.759 & 0.760 \\
\hline
Recall & 0.634 & 0.632 & 0.635 & 0.644 & 0.627 & 0.634 & 0.644 & 0.649 & 0.667 \\
\hline
Loss & 0.299 & 0.298 & 0.298 & 0.300 & 0.301 & 0.298 & 0.301 & 0.297 & 0.310 \\
\hline
True Positive & 1306 & 1315 & 1328 & 1306 & 1292 & 1307 & 1327 & 1338 & 1375 \\
\hline
True Negative & 6251 & 6244 & 6237 & 6251 & 6260 & 6250 & 6238 & 6235 & 6181 \\
\hline
False Positive & 112 & 119 & 126 & 112 & 103 & 113 & 125 & 128 & 182 \\
\hline
False Negative & 753 & 744 & 731 & 753 & 767 & 752 & 732 & 721 & 684 \\
\hline

\end{tabular}
}
\caption{Dataset 2}
\end{table}

\clearpage

\clearpage

\begin{samepage}

\begin{table}[H]
\centering
\resizebox{\textwidth}{!}{
\begin{tabular}{|l|c|c|c|c|c|c|c|c|c|}
\hline
\multirow{2}{*}{Metrics}
& \multicolumn{3}{|c|}{Single LSTM}
& \multicolumn{3}{|c|}{Stacked LSTM}
& \multicolumn{3}{|c|}{Stacked LSTM with Dense layers} \\
\cline{2-10}
& Model 1 & Model 2 & Model 3
& Model 1 & Model 2 & Model 3
& Model 1 & Model 2 & Model 3 \\
\hline
Accuracy & 0.896 & 0.896 & 0.896 & 0.894 & 0.896 & 0.896 & 0.894 & 0.893 & 0.896 \\
\hline
Precision & 0.890 & 0.895 & 0.895 & 0.931 & 0.895 & 0.945 & 0.902 & 0.943 & 0.891 \\
\hline
F1 score & 0.758 & 0.759 & 0.759 & 0.753 & 0.759 & 0.753 & 0.750 & 0.753 & 0.759 \\
\hline
Recall & 0.661 & 0.659 & 0.659 & 0.632 & 0.659 & 0.626 & 0.642 & 0.627 & 0.661 \\
\hline
Loss & 0.306 & 0.306 & 0.305 & 0.306 & 0.305 & 0.305 & 0.307 & 0.307 & 0.305 \\
\hline
True Positive & 1441 & 1437 & 1437 & 1400 & 1437 & 1437 & 1400 & 1373 & 1441 \\
\hline
True Negative & 6464 & 6474 & 6474 & 6489 & 6474 & 6474 & 6489 & 6510 & 6465 \\
\hline
False Positive & 177 & 167 & 167 & 152 & 167 & 167 & 152 & 131 & 176 \\
\hline
False Negative & 739 & 743 & 743 & 780 & 743 & 743 & 780 & 807 & 739 \\
\hline
\end{tabular}
}
\caption{Dataset 3}
\end{table}

\begin{table}[H]
\centering
\resizebox{\textwidth}{!}{
\begin{tabular}{|l|c|c|c|c|c|c|c|c|c|}
\hline
\multirow{2}{*}{Metrics}
& \multicolumn{3}{|c|}{Single LSTM}
& \multicolumn{3}{|c|}{Stacked LSTM}
& \multicolumn{3}{|c|}{Stacked LSTM with Dense layers} \\
\cline{2-10}
& Model 1 & Model 2 & Model 3
& Model 1 & Model 2 & Model 3
& Model 1 & Model 2 & Model 3 \\
\hline
Accuracy & 0.905 & 0.905 & 0.906 & 0.905 & 0.905 & 0.906 & 0.906 & 0.906 & 0.906 \\
\hline
Precision & 0.900 & 0.900 & 0.900 & 0.900 & 0.900 & 0.900 & 0.900 & 0.900 & 0.900 \\
\hline
F1 score & 0.767 & 0.767 & 0.767 & 0.767 & 0.767 & 0.767 & 0.767 & 0.767 & 0.767 \\
\hline
Recall & 0.668 & 0.668 & 0.668 & 0.668 & 0.668 & 0.668 & 0.668 & 0.668 & 0.668 \\
\hline
Loss & 0.305 & 0.304 & 0.304 & 0.304 & 0.305 & 0.304 & 0.305 & 0.304 & 0.306 \\
\hline
True Positive & 1699 & 1699 & 1700 & 1699 & 1699 & 1700 & 1700 & 1700 & 1700 \\
\hline
True Negative & 8223 & 8223 & 8223 & 8223 & 8223 & 8223 & 8223 & 8223 & 8223 \\
\hline
False Positive & 187 & 187 & 187 & 187 & 187 & 187 & 187 & 187 & 187 \\
\hline
False Negative & 843 & 843 & 842 & 843 & 843 & 842 & 842 & 842 & 842 \\
\hline
\end{tabular}
}
\caption{Dataset 4}
\end{table}

\begin{table}[H]
\centering
\resizebox{\textwidth}{!}{
\begin{tabular}{|l|c|c|c|c|c|c|c|c|c|}
\hline
\multirow{2}{*}{Metrics}
& \multicolumn{3}{|c|}{Single LSTM}
& \multicolumn{3}{|c|}{Stacked LSTM}
& \multicolumn{3}{|c|}{Stacked LSTM with Dense layers} \\
\cline{2-10}
& Model 1 & Model 2 & Model 3
& Model 1 & Model 2 & Model 3
& Model 1 & Model 2 & Model 3 \\
\hline
Accuracy & 0.887 & 0.892 & 0.892 & 0.892 & 0.887 & 0.892 & 0.892 & 0.892 & 0.892 \\
\hline
Precision & 0.812 & 0.892 & 0.892 & 0.892 & 0.908 & 0.900 & 0.892 & 0.892 & 0.892 \\
\hline
F1 score & 0.736 & 0.726 & 0.726 & 0.726 & 0.767 & 0.767 & 0.726 & 0.726 & 0.726 \\
\hline
Recall & 0.673 & 0.612 & 0.612 & 0.612 & 0.668 & 0.668 & 0.612 & 0.612 & 0.612 \\
\hline
Loss & 0.327 & 0.327 & 0.327 & 0.327 & 0.327 & 0.327 & 0.327 & 0.327 & 0.327 \\
\hline
True Positive & 1730 & 1575 & 1575 & 1575 & 1730 & 1575 & 1575 & 1575 & 1575 \\
\hline
True Negative & 8006 & 8215 & 8215 & 8215 & 8006 & 8215 & 8215 & 8215 & 8215 \\
\hline
False Positive & 399 & 190 & 190 & 190 & 399 & 190 & 190 & 190 & 190 \\
\hline
False Negative & 840 & 995 & 995 & 995 & 840 & 995 & 995 & 995 & 995 \\
\hline
\end{tabular}
}
\caption{Dataset 5}
\end{table}

\end{samepage}

\begin{table}[H]
\centering
\resizebox{\textwidth}{!}{
\begin{tabular}{|l|c|c|c|c|c|c|c|c|c|}
\hline
\multirow{2}{*}{Metrics}
& \multicolumn{3}{|c|}{Single LSTM}
& \multicolumn{3}{|c|}{Stacked LSTM}
& \multicolumn{3}{|c|}{Stacked LSTM with Dense layers} \\
\cline{2-10}
& Model 1 & Model 2 & Model 3
& Model 1 & Model 2 & Model 3
& Model 1 & Model 2 & Model 3 \\
\hline
Accuracy & 0.766 & 0.766 & 0.766 & 0.766 & 0.766 & 0.766 & 0.766 & 0.766 & 0.766 \\
\hline
Precision & 0 & 0 & 0 & 0 & 0 & 0 & 0 & 0 & 0 \\
\hline
F1 score & 0 & 0 & 0 & 0 & 0 & 0 & 0 & 0 & 0 \\
\hline
Recall & 0 & 0 & 0 & 0 & 0 & 0 & 0 & 0 & 0 \\
\hline
Loss & 0.538 & 0.538 & 0.540 & 0.543 & 0.543 & 0.539 & 0.543 & 0.543 & 0.543 \\
\hline
True Positive & 0 & 0 & 0 & 0 & 0 & 0 & 0 & 0 & 0 \\
\hline
True Negative & 8418 & 8418 & 8418 & 8418 & 8418 & 8418 & 8418 & 8418 & 8418 \\
\hline
False Positive & 0 & 0 & 0 & 0 & 0 & 0 & 0 & 0 & 0 \\
\hline
False Negative & 2560 & 2560 & 2560 & 2560 & 2560 & 2560 & 2560 & 2560 & 2560 \\
\hline
\end{tabular}
}
\caption{Dataset 6}
\end{table}

\begin{table}[h!]
\centering
\begin{tabular}{|l|c|}
\hline
\textbf{Metrics} & \textbf{Decision Tree Model} \\ \hline
Accuracy & 0.962 \\ \hline
Precision & 0.914 \\ \hline
F1 score & 0.920 \\ \hline
Recall & 0.925 \\ \hline
Loss & -- \\ \hline
True Positive & 6050 \\ \hline
True Negative & 20542 \\ \hline
False Positive & 565 \\ \hline
False Negative & 484 \\ \hline
\end{tabular}
\caption{Performance Metrics of Decision Tree Model}
\label{tab:dt_metrics}
\end{table}

In this study, we evaluated the effectiveness of time series models (LSTM variants) against a traditional single-shot model (Decision Tree) for cheat detection in FPS games using game tick sequence data. The experiments were conducted across 6 dataset configurations with varying temporal context windows, and 9 LSTM model configurations per dataset.
The most critical metric for this domain is the \textbf{False Positive Rate (FPR = FP / (FP + TN))}, as falsely accusing an innocent player of cheating is a severe and often irreversible action. The Decision Tree baseline achieves:

{\small
\[
\text{DT FPR} = \frac{565}{565 + 20542} = \frac{565}{21107} = 2.68\%
\]
}

Across the LSTM experiments, the best FPR per dataset (taken as the model with the lowest FP count) is as follows:

\begin{table}[h]
\centering
\label{tab:fpr_window}
\begin{tabular}{|l|c|c|c|c|c|}
\hline
\textbf{Dataset} & \textbf{Max Window (ticks)} & \textbf{FP} & \textbf{TN} & \textbf{FP / (FP + TN)} & \textbf{FPR} \\
\hline
Dataset 1 & 128 & 50  & 5092 & $\frac{50}{50 + 5092}$   & 0.97\% \\[6pt]
\hline
Dataset 2 & 96  & 103 & 6260 & $\frac{103}{103 + 6260}$ & 1.62\% \\[6pt]
\hline
Dataset 3 & 64  & 131 & 6510 & $\frac{131}{131 + 6510}$ & 1.97\% \\[6pt]
\hline
Dataset 4 & 32  & 187 & 8223 & $\frac{187}{187 + 8223}$ & 2.22\% \\[6pt]
\hline
Dataset 5 & 16  & 190 & 8215 & $\frac{190}{190 + 8215}$ & 2.26\% \\[6pt]
\hline
Dataset 6 & 1   & 0   & 8418 & $\frac{0}{0 + 8418}$     & 0.00\% \\
\hline
\end{tabular}
\caption{False Positive Rate of best performing models across datasets}
\end{table}

This reveals a clear and monotonic relationship: \textbf{larger temporal context windows consistently produce lower false positive rates}, with Dataset 1's best model (Stacked LSTM with Dense layers, Model 1) achieving an FPR of 0.97\%, less than half the Decision Tree's FPR of 2.68\%.

Dataset 6, where the context window is reduced to a single tick, results in complete model collapse across all 9 LSTM configurations. Every model defaults to predicting the majority class (non-cheating), yielding zero true positives and an accuracy of:

{\small
\[
\text{Majority class accuracy} = \frac{8418}{8418 + 2560} = \frac{8418}{10978} = \textbf{76.7}\%
\]
}

This confirms that temporal context is the essential signal for this classification task. Without it, even complex neural architectures fail entirely.
Regarding model architecture, increasing LSTM units from 32 to 64 to 128 provides negligible improvement. For example in Dataset 4, accuracy across all 9 models ranges only from 0.905 to 0.906, a difference of \textbf{0.001 (0.1\%)}. This suggests the bottleneck is the sequence construction (window size and tick context) rather than model capacity, and that simpler LSTM configurations are preferable given their lower computational cost.

\par The Decision Tree model does achieve a notably higher recall of 0.925, compared to the best LSTM recall of approximately 0.673 (Dataset 5, Single LSTM Model 1): 

\begin{table}[h!]
\centering
\begin{tabular}{|l|c|c|}
\hline
\textbf{Model} & \textbf{TP / (TP + FN)} & \textbf{Recall} \\ \hline
DT Recall   & $\frac{6050}{6050 + 484}$ & 92.6\% \\[6pt] \hline
Best LSTM Recall & $\frac{1730}{1730 + 840}$ & 67.3\% \\[6pt] \hline
\end{tabular}
\caption{Recall on decision tree and LSTM models}
\label{tab:recall_dt_lstm}
\end{table}

\vspace{9cm}

\par This means the Decision Tree catches more cheaters, but at the cost of a higher false accusation rate. In the context of FPS anti-cheat systems, we argue that missing a cheater is a less severe error than wrongly punishing an innocent player a missed cheater may be detected in future matches, whereas a wrongful ban directly harms a legitimate player's experience. This domain-driven priority justifies favouring the lower FPR of the LSTM models despite the recall tradeoff.

In conclusion, the best overall configuration is the \textbf{Stacked LSTM with Dense layers (Model 1) trained on Dataset 1} (Tick Delta Negative=56, Tick Delta Positive=24, Max Fight Size=128), achieving an accuracy of \textbf{88.6\%}, with an FPR of \textbf{0.97\%}, outperforming the Decision Tree baseline FPR of \textbf{2.68\%} by a factor of \textbf{2.76x}.

Future work could explore attention-based sequence models (such as Transformers) or hybrid architectures that may further improve recall without sacrificing the low false positive rate achieved here.

\newpage

\section{Conclusion}

\par  In conclusion, this research demonstrates that deep learning-based sequence models offer a meaningful advantage over traditional single-shot classification approaches for aimbot detection in FPS games. Through systematic experimentation across 6 dataset configurations and 9 LSTM model variants, we show that temporal context is the defining factor in minimising false positive rates, with the best LSTM configuration achieving an FPR of 0.97\%, 2.76x lower than the Decision Tree baseline of 2.68\%. While the Decision Tree achieves a higher accuracy of 96.2\% and recall of 92.6\%, its higher false positive rate makes it less suitable for anti-cheat systems where wrongly penalising an innocent player carries significant consequences. The complete failure of all models on Dataset 6 (1-tick window) further confirms that without sufficient temporal context, even complex architectures cannot distinguish cheating behaviour from legitimate gameplay. Future work could explore attention-based architectures such as Transformers, or hybrid models that retain the low false positive rate of LSTM while closing the recall gap against the Decision Tree baseline. Additionally, a sequential ensemble approach - first applying the Decision Tree for its high recall of 92.6\% to flag potential cheaters, followed by the LSTM model to filter out false accusations with its lower FPR of 0.97\% could offer the best of both models, maintaining strong cheat detection while minimising wrongful penalties.

\end{document}